\documentclass[10pt,twocolumn,letterpaper]{article}

\usepackage{iccv}
\usepackage{times}
\usepackage{epsfig}
\usepackage{graphicx}
\usepackage{amsmath}
\usepackage{amssymb}

\usepackage{booktabs}
\usepackage{enumitem}

\usepackage[pagebackref=true,breaklinks=true,letterpaper=true,colorlinks,bookmarks=false]{hyperref}

\iccvfinalcopy 

\def\iccvPaperID{1096} 

\ificcvfinal\fi
\begin{document}
	
	\title{Ego-Pose Estimation and Forecasting as Real-Time PD Control}
	
	\author{Ye Yuan \quad Kris Kitani\\
		Carnegie Mellon University\\
		{\tt\small \{yyuan2, kkitani\}@cs.cmu.edu}
	}
	
	\maketitle

	\begin{abstract}\vspace{-3mm}
		We propose the use of a proportional-derivative (PD) control based policy learned via reinforcement learning (RL) to estimate and forecast 3D human pose from egocentric videos. The method learns directly from unsegmented egocentric videos and motion capture data consisting of various complex human motions (e.g., crouching, hopping, bending, and motion transitions). We propose a video-conditioned recurrent control technique to forecast physically-valid and stable future motions of arbitrary length. We also introduce a value function based fail-safe mechanism which enables our method to run as a single pass algorithm over the video data. Experiments with both controlled and in-the-wild data show that our approach outperforms previous art in both quantitative metrics and visual quality of the motions, and is also robust enough to transfer directly to real-world scenarios. Additionally, our time analysis shows that the combined use of our pose estimation and forecasting can run at 30 FPS, making it suitable for real-time applications.\footnote{Project page: \href{https://www.ye-yuan.com/ego-pose}{https://www.ye-yuan.com/ego-pose}}
	\end{abstract}
	
	\vspace{-5mm}
	
	\section{Introduction}
	
	With a single wearable camera, our goal is to estimate and forecast a person's pose sequence for a variety of complex motions. Estimating and forecasting complex human motions with egocentric cameras can be the cornerstone of many useful applications. In medical monitoring, the inferred motions can help physicians remotely diagnose patients' condition in motor rehabilitation. In virtual or augmented reality, anticipating motions can help allocate limited computational resources to provide better responsiveness. For athletes, the forecasted motions can be integrated into a coaching system to offer live feedback and reinforce good movements. In all these applications, human motions are very complex, as periodical motions (\eg, walking, running) are often mixed with non-periodical motions (\eg, turning, bending, crouching). It is challenging to estimate and forecast such complex human motions from egocentric videos due to the multi-modal nature of the data.
	
	It has been shown that if the task of pose estimation can be limited to a single mode of action such as running or walking, it is possible to estimate a physically-valid pose sequence. Recent work by Yuan and Kitani~\cite{yuan20183d} has formulated egocentric pose estimation as a Markov decision process (MDP): a humanoid agent driven by a control policy with visual input to generate a pose sequence inside a physics simulator. They use generative adversarial imitation learning (GAIL~\cite{ho2016generative}) to solve for the optimal control policy. By design, this approach guarantees that the estimated pose sequence is physically-valid. However, their method focuses on a single action modality (\ie, simple periodical motions including walking and running). The approach also requires careful segmentation of the demonstrated motions, due to the instability of adversarial training when the data is multi-modal. To address these issues, we propose an ego-pose estimation approach that can learn a motion policy directly from unsegmented multi-modal motion demonstrations.
	
	Unlike the history of work on egocentric pose estimation, there has been no prior work addressing the task of egocentric pose forecasting. Existing works on 3D pose forecasting not based on egocentric sensing take a pose sequence as input and uses recurrent models to output a future pose sequence by design~\cite{fragkiadaki2015recurrent, jain2016structural, butepage2017deep, li2017auto}. Even with the use of a 3D pose sequence as a direct input, these methods tend to produce unrealistic motions due to error accumulation (covariate shift~\cite{quionero2009dataset}) caused by feeding predicted pose back to the network without corrective interaction with the learning environment. More importantly, these approaches often generate physically-invalid pose sequences as they are trained only to mimic motion kinematics, disregarding causal forces like the laws of physics or actuation constraints. In this work, we propose a method that directly takes noisy observations of past egocentric video as input to forecast stable and physically-valid future human motions.
	
	We formulate both egocentric pose estimation and forecasting as a MDP. The humanoid control policy takes as input the current state of the humanoid for both inference tasks. Additionally, the visual context from the entire video is used as input for the pose estimation task. In the case of the forecasting task, only the visual input observed up to the current time step is used. For the action space of the policy, we use target joint positions of proportional-derivative (PD) controllers~\cite{tan2011stable} instead of direct joint torques. The PD controllers act like damped springs and compute the torques to be applied at each joint. This type of action design is more capable of actuating the humanoid to perform highly dynamic motions~\cite{peng2018deepmimic}. As deep reinforcement learning (DeepRL) based approaches for motion imitation~\cite{peng2018deepmimic, peng2018sfv} have proven to be more robust than GAIL based methods~\cite{yuan20183d, merel2017learning, wang2017robust}, we utilize DeepRL to encourage the motions generated by the control policy to match the ground-truth. However, reward functions designed for motion imitation methods are not suited for our task because they are tailored to learning locomotions from short segmented motion clips, while our goal is to learn to estimate and forecast complex human motions from unsegmented multi-modal motion data. Thus, we propose a new reward function that is specifically designed for this type of data. For forecasting, we further employ a decaying reward function to focus on forecasting for frames in the near future. Since we only take past video frames as input and the video context is fixed during forecasting, we use a recurrent control policy to better encode the phase of the human motion.
	
	A unique problem encountered by the control-based approach taken in this work is that the humanoid being actuated in the physics simulator can fall down. Specifically, extreme domain shifts in the visual input at test time can cause irregular control actions. As a result, this irregularity in control actions causes the humanoid to lose balance and fall in the physics environment, preventing the method from providing any pose estimates. The control-based method proposed in~\cite{yuan20183d} prevented falling by fine-tuning the policy at test time as a batch process. As a result, this prohibits its use in streaming or real-time applications. Without fine-tuning, their approach requires that we reset the humanoid state to some reasonable starting state to keep producing meaningful pose estimates. However, it is not clear when to re-estimate the state. To address this issue of the humanoid falling in the physics simulator at test time, we propose a fail-safe mechanism based on a value function estimate used in the policy gradient method. The mechanism can anticipate falling much earlier and stabilize the humanoid before producing bad pose estimates.
	
	We validate our approach for egocentric pose estimation and forecasting on a large motion capture (MoCap) dataset and an in-the-wild dataset consisting of various human motions (jogging, bending, crouching, turning, hopping, leaning, motion transitions, \etc). Experiments on pose estimation show that our method can learn directly from unsegmented data and outperforms state-of-the-art methods in terms of both quantitative metrics and visual quality of the motions. Experiments on pose forecasting show that our approach can generate intuitive future motions and is also more accurate compared to other baselines. Our in-the-wild experiments show that our method transfers well to real-world scenarios without the need for any fine-tuning. Our time analysis show that our approach can run at 30 FPS, making it suitable for many real-time applications.
	
	In summary, our contributions are as follows: (1) We propose a DeepRL-based method for egocentric pose estimation that can learn from unsegmented MoCap data and estimate accurate and physically-valid pose sequences for complex human motions. (2) We are the first to tackle the problem of egocentric pose forecasting and show that our method can generate accurate and stable future motions. (3) We propose a fail-safe mechanism that can detect instability of the humanoid control policy, which prevents generating bad pose estimates. (4) Our model trained with MoCap data transfers well to real-world environments without any fine-tuning. (5) Our time analysis show that our pose estimation and forecasting algorithms can run in real-time.

	\section{Related Work}
	\vspace{1mm}
	\noindent\textbf{3D Human Pose Estimation.} Third-person pose estimation has long been studied by the vision community~\cite{liu2015survey, sarafianos20163d}. Existing work leverages the fact that the human body is visible from the camera. Traditional methods tackle the depth ambiguity with strong priors such as shape models~\cite{zhou2017sparse, bogo2016keep}. Deep learning based approaches~\cite{zhou2016sparseness, pavlakos2017coarse, mehta2017vnect, tung2017self} have also succeeded in directly regressing images to 3D joint locations with the help of large-scale MoCap datasets~\cite{ionescu2014human3}. To achieve better performance for in-the-wild images, weakly-supervised methods~\cite{zhou2017towards, rogez2016mocap, kanazawa2018end} have been proposed to learn from images without annotations. Although many of the state-of-art approaches predict pose for each frame independently, several works have utilized video sequences to improve temporal consistency~\cite{tekin2016direct, xu2018monoperfcap, dabral2017structure, kanazawa2018learn}.
	
	Limited amount of research has looked into egocentric pose estimation. Most existing methods only estimate the pose of visible body parts~\cite{li2013model,li2013pixel,ren2010figure,arikan2003motion,rogez2015first}. Other approaches utilize 16 or more body-mounted cameras to infer joint locations via structure from motion~\cite{shiratori2011motion}. Specially designed head-mounted rigs have been used for markerless motion capture~\cite{rhodin2016egocap, xu2019mo, tome2019xr}, where~\cite{tome2019xr} utilizes photorealistic synthetic data. Conditional random field based methods~\cite{jiang2017seeing} have also been proposed to estimate a person's full-body pose with a wearable camera. The work most related to ours is~\cite{yuan20183d} which formulates egocentric pose estimation as a Markov decision process to enforce physics constraints and solves it by adversarial imitation learning. It shows good results on simple periodical human motions but fails to estimate complex non-periodical motions. Furthermore, they need fine-tuning at test time to prevent the humanoid from falling. In contrast, we propose an approach that can learn from unsegmented MoCap data and estimate various complex human motions in real-time without fine-tuning.
	
	\vspace{1mm}
	\noindent\textbf{Human Motion Forecasting.}
	Plenty of work has investigated third-person~\cite{Xie2013InferringM, ma2017forecasting, ballan2016knowledge, kitani2012activity, alahi2016social, robicquet2016learning, yagi2018future} and first-person~\cite{soo2016egocentric} trajectory forecasting, but this line of work only forecasts a person's future positions instead of poses. There are also works focusing on predicting future motions in image space~\cite{finn2016unsupervised, walker2016uncertain, walker2017pose, denton2017unsupervised, xue2016visual, li2018flow, gao2018im2flow}. Other methods use past 3D human pose sequence as input to predict future human motions~\cite{fragkiadaki2015recurrent, jain2016structural, butepage2017deep, li2017auto}. Recently,~\cite{kanazawa2018learn, chao2017forecasting} forecast a person's future 3D poses from third-person static images, which require the person to be visible. Different from previous work, we propose to forecast future human motions from egocentric videos where the person can hardly be seen.
	
	\vspace{1mm}
	\noindent\textbf{Humanoid Control from Imitation.} The idea of using reference motions has existed for a long time in computer animation. Early work has applied this idea to bipedal locomotions with planar characters~\cite{sharon2005synthesis, sok2007simulating}. Model-based methods~\cite{yin2007simbicon, muico2009contact, lee2010data} generate locomotions with 3D humanoid characters by tracking reference motions. Sampling-based control methods~\cite{liu2010sampling, liu2016guided, liu2017learning} have also shown great success in generating highly dynamic humanoid motions. DeepRL based approaches have utilized reference motions to shape the reward function~\cite{peng2017deeploco, peng2017learning}. Approaches based on GAIL~\cite{ho2016generative} have also been proposed to eliminate the need for manual reward engineering~\cite{merel2017learning, wang2017robust, yuan20183d}. The work most relevant to ours is DeepMimic~\cite{peng2018deepmimic} and its video variant~\cite{peng2018sfv}. DeepMimic has shown beautiful results on human locomotion skills with manually designed reward and is able to combine learned skills to achieve different tasks. However, it is only able to learn skills from segmented motion clips and relies on the phase of motion as input to the policy. In contrast, our approach can learn from unsegmented MoCap data and use the visual context as a natural alternative to the phase variable.

	\section{Methodology}
	\begin{figure*}
		\centering
		\includegraphics[width=\textwidth]{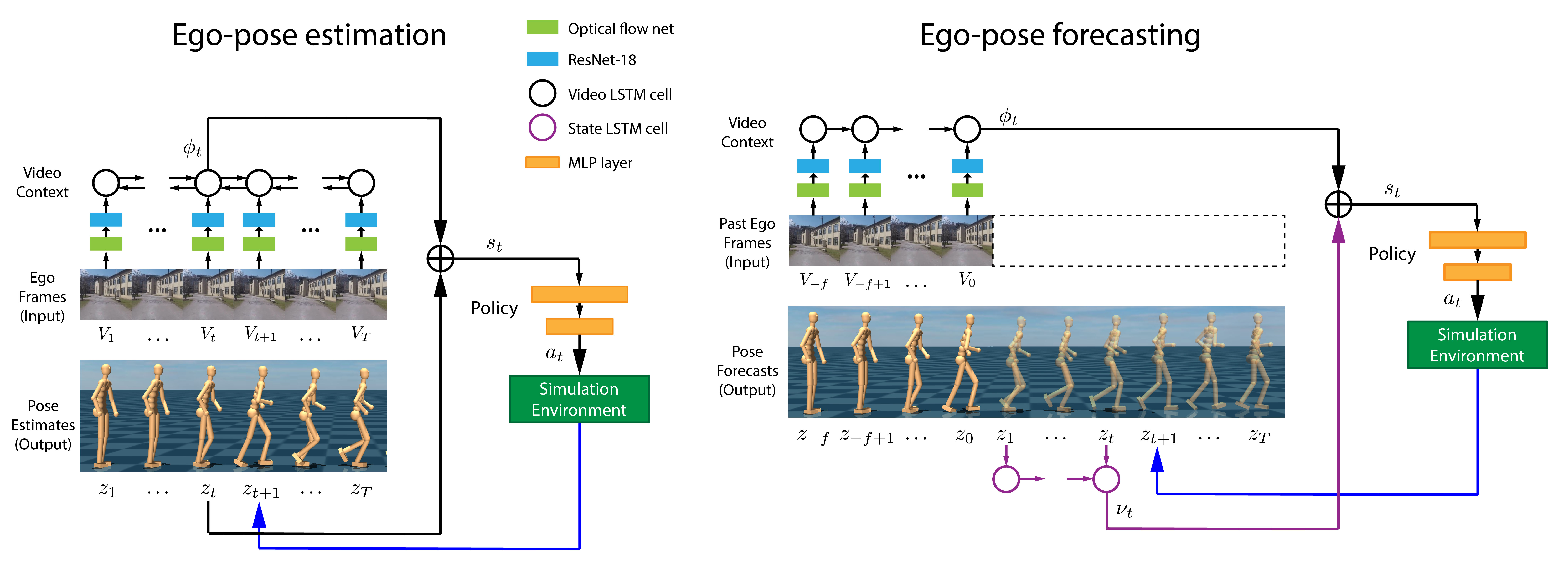}
		\caption{Overview for ego-pose estimation and forecasting. The policy takes in the humanoid state $z_t$ (estimation) or recurrent state feature $\nu_t$ (forecasting) and the visual context $\phi_t$ to output the action $a_t$, which generates the next humanoid state $z_{t+1}$ through physics simulation. \textbf{Left:} For ego-pose estimation, the visual context $\phi_t$ is computed from the entire video $V_{1:T}$ using a Bi-LSTM to encode CNN features. \textbf{Right:} For ego-pose forecasting, $\phi_t$ is computed from past frames $V_{-f:0}$ using a forward LSTM and is kept fixed for all $t$.}
		\label{fig:overview}
		\vspace{-5mm}
	\end{figure*}
	
	We choose to model human motion as the result of the optimal control of a dynamical system governed by a cost (reward) function, as control theory provides mathematical machinery necessary to explain human motion under the laws of physics. In particular, we use the formalism of the Markov Decision process (MDP). The MDP is defined by a tuple $\mathcal{M} = \left(S, A, P, R, \gamma\right)$ of states, actions, transition dynamics, a reward function, and a discount factor. 
	
	\vspace{1mm}
	\noindent\textbf{State.} The state $s_t$ consists of both the state of the humanoid $z_t$ and the visual context $\phi_t$. The humanoid state $z_t$ consists of the pose $p_t$ (position and orientation of the root, and joint angles) and velocity $v_t$ (linear and angular velocities of the root, and joint velocities). All features are computed in the humanoid's local heading coordinate frame which is aligned with the root link's facing direction. The visual context $\phi_t$ varies depending on the task (pose estimation or forecasting) which we will address in Sec.~\ref{sec:estimation} and~\ref{sec:forecast}.
	
	\vspace{1mm}
	\noindent\textbf{Action.} The action $a_t$ specifies the target joint angles for the Proportional-Derivative (PD) controller at each degree of freedom (DoF) of the humanoid joints except for the root. For joint DoF $i$, the torque to be applied is computed as
	\begin{equation}
	\tau^{i} = k_p^i(a_t^i - p_t^i) - k_d^i v_t^i\,,
	\end{equation}
	where $k_p$ and $k_d$ are manually-specified gains.
	Our policy is queried at $30$Hz while the simulation is running at $450$Hz, which gives the PD-controllers 15 iterations to try to reach the target positions. Compared to directly using joint torques as the action, this type of action design increases the humanoid's capability of performing highly dynamic motions~\cite{peng2018deepmimic}.
	
	\vspace{1mm}
	\noindent\textbf{Policy.} The policy $\pi_\theta(a_t|s_t)$ is represented by a Gaussian distribution with a fixed diagonal covariance matrix $\Sigma$. We use a neural network with parameter $\theta$ to map state $s_t$ to the mean $\mu_t$ of the distribution. We use a multilayer perceptron (MLP) with two hidden layers $(300, 200)$ and ReLU activation to model the network. Note that at test time we always choose the mean action from the policy to prevent performance drop from the exploration noise.

	\vspace{1mm}
	\noindent\textbf{Solving the MDP.} At each time step, the humanoid agent in state $s_t$ takes an action $a_t$ sampled from a policy $\pi(a_t|s_t)$, and the environment generates the next state $s_{t+1}$ through physics simulation and gives the agent a reward $r_t$ based on how well the humanoid motion aligns with the ground-truth. This process repeats until some termination condition is triggered such as when the time horizon is reached or the humanoid falls. To solve this MDP, we apply policy gradient methods (\eg, PPO~\cite{schulman2017proximal}) to obtain the optimal policy $\pi^\star$ that maximizes the expected discounted return $\mathbb{E}\left[\sum_{t=1}^T \gamma^{t-1} r_t\right]$. At test time, starting from some initial state $s_1$, we rollout the policy $\pi^\star$ to generate state sequence $s_{1:T}$, from which we extract the output pose sequence $p_{1:T}$.

	\subsection{Ego-pose Estimation}
	\label{sec:estimation}
	The goal of egocentric pose estimation is to use video frames $V_{1:T}$ from a wearable camera to estimate the person's pose sequence $p_{1:T}$. To learn the humanoid control policy $\pi(a_t|z_t,\phi_t)$ for this task, we need to define the procedure for computing the visual context $\phi_t$ and the reward function $r_t$. As shown in Fig.~\ref{fig:overview} (Left), the visual context $\phi_t$ is computed from the video $V_{1:T}$. Specifically, we calculate the optical flow for each frame and pass it through a CNN to extract visual features $\psi_{1:T}$. Then we feed $\psi_{1:T}$ to a bi-directional LSTM to generate the visual context $\phi_{1:T}$, from which we obtain per frame context $\phi_t$. For the starting state $z_1$, we set it to the ground-truth $\hat{z}_1$ during training. To encourage the pose sequence $p_{1:T}$ output by the policy to match the ground-truth $\hat{p}_{1:T}$, we define our reward function as
	\begin{equation}
	\label{eq:reward}
	r_t = w_q r_q + w_e r_e + w_p r_p + w_v r_v\,,
	\end{equation}
	where $w_q, w_e, w_p, w_v$ are weighting factors.
	
	The pose reward $r_q$ measures the difference between pose $p_t$ and the ground-truth $\hat{p}_t$ for non-root joints. We use $q_t^j$ and $\hat{q}_t^j$ to denote the local orientation quaternion of joint $j$ computed from $p_t$ and $\hat{p}_t$ respectively. We use $q_1 \ominus q_2$ to denote the relative quaternion from $q_2$ to $q_1$, and $\|q\|$ to compute the rotation angle of $q$.
	\begin{equation}
	r_q = \exp\left[-2\left(\sum_{j}\|q_t^j\ominus \hat{q}_t^j\|^2\right)\right]\,.
	\end{equation}
	The end-effector reward $r_e$ evaluates the difference between local end-effector vector $e_t$ and the ground-truth $\hat{e}_t$. For each end-effector $e$ (feet, hands, head), $e_t$ is computed as the vector from the root to the end-effector.
	\begin{equation}
	r_e = \exp\left[-20\left(\sum_{e}\|e_t- \hat{e}_t\|^2\right)\right]\,.
	\end{equation}
	The root pose reward $r_p$ encourages the humanoid's root joint to have the same height $h_t$ and orientation quaternion $q_t^r$ as the ground-truth $\hat{h}_t$ and $\hat{q}_t^r$.
	\begin{equation}
	r_p = \exp\left[-300\left((h_t - \hat{h}_t)^2 + \|q_t^r\ominus \hat{q}_t^r\|^2\right) \right]\,.
	\end{equation}
	The root velocity reward $r_v$ penalizes the deviation of the root's linear velocity $l_t$ and angular velocity $\omega_t$ from the ground-truth $\hat{l}_t$ and $\hat{\omega}_t$. The ground-truth velocities can be computed by the finite difference method.
	\begin{equation}
	r_v = \exp\left[-\|l_t- \hat{l}_t\|^2 -0.1 \|\omega_t^r-\hat{\omega}_t^r\|^2 \right]\,.
	\end{equation}
	Note that all features are computed inside the local heading coordinate frame instead of the world coordinate frame, which is crucial to learn from unsegmented MoCap data for the following reason: when imitating an unsegmented motion demonstration, the humanoid will drift from the ground-truth motions in terms of global position and orientation because the errors made by the policy accumulate; if the features are computed in the world coordinate, their distance to the ground-truth quickly becomes large and the reward drops to zero and stops providing useful learning signals. Using local features ensures that the reward is well-shaped even with large drift. To learn global motions such as turning with local features, we use the reward $r_v$ to encourage the humanoid's root to have the same linear and angular velocities as the ground-truth.
	
	\vspace{1mm}\noindent\textbf{Initial State Estimation.} As we have no access to the ground-truth humanoid starting state $z_1$ at test time, we need to learn a regressor $\mathcal{F}$ that maps video frames $V_{1:T}$ to their corresponding state sequence $z_{1:T}$. $\mathcal{F}$ uses the same network architecture as ego-pose estimation (Fig.~\ref{fig:overview} (Left)) for computing the visual context $\phi_{1:T}$ . We then pass $\phi_{1:T}$ through an MLP with two hidden layers (300, 200) to output the states. We use the mean squared error (MSE) as the loss function: $L(\zeta) = \frac{1}{T}\sum_{t=1}^T \|\mathcal{F}(V_{1:T})_t - z_t\|^2$, where $\zeta$ is the parameters of $\mathcal{F}$. The optimal $\mathcal{F}^\star$ can be obtained by an SGD-based method.

	\subsection{Ego-pose Forecasting}
	\label{sec:forecast}
	For egocentric pose forecasting, we aim to use past video frames $V_{-f:0}$ from a wearable camera to forecast the future pose sequence $p_{1:T}$ of the camera wearer. We start by defining the visual context $\phi_t$ used in the control policy $\pi$. As shown in Fig.~\ref{fig:overview} (Right), the visual context $\phi_t$ for this task is computed from past frames $V_{-f:0}$ and is kept fixed for all time $t$ during a policy rollout.  We compute the optical flow for each frame and use a CNN to extract visual features $\psi_{-f:0}$. We then use a forward LSTM to summarize $\psi_{-f:0}$ into the visual context $\phi_t$. For the humanoid starting state $z_1$, we set it to the ground-truth $\hat{z}_1$, which at test time is provided by ego-pose estimation on $V_{-f:0}$. Now we define the reward function for the forecasting task. Due to the stochasticity of human motions, the same past frames can correspond to multiple future pose sequences. As the time step $t$ progresses, the correlation between pose $p_t$ and past frames $V_{-f:0}$ diminishes. This motivates us to use a reward function that focuses on frames closer to the starting frame:
	\begin{equation}
	\tilde{r}_t = \beta r_t\,,
	\end{equation}
	where $\beta = (T - t) / T$ is a linear decay factor and $r_t$ is defined in  Eq.~\ref{eq:reward}. Unlike ego-pose estimation, we do not have new video frame coming as input for each time step $t$, which can lead to ambiguity about the motion phase, such as whether the human is standing up or crouching down. To better encode the phase of human motions, we use a recurrent policy $\pi(a_t|\nu_t, \phi_t)$ where $\nu_t \in \mathbb{R}^{128}$ is the output of a forward LSTM encoding the state forecasts $z_{1:t}$ so far.

	\subsection{Fail-safe Mechanism}
	\begin{figure}
		\centering
		\includegraphics[width=\linewidth]{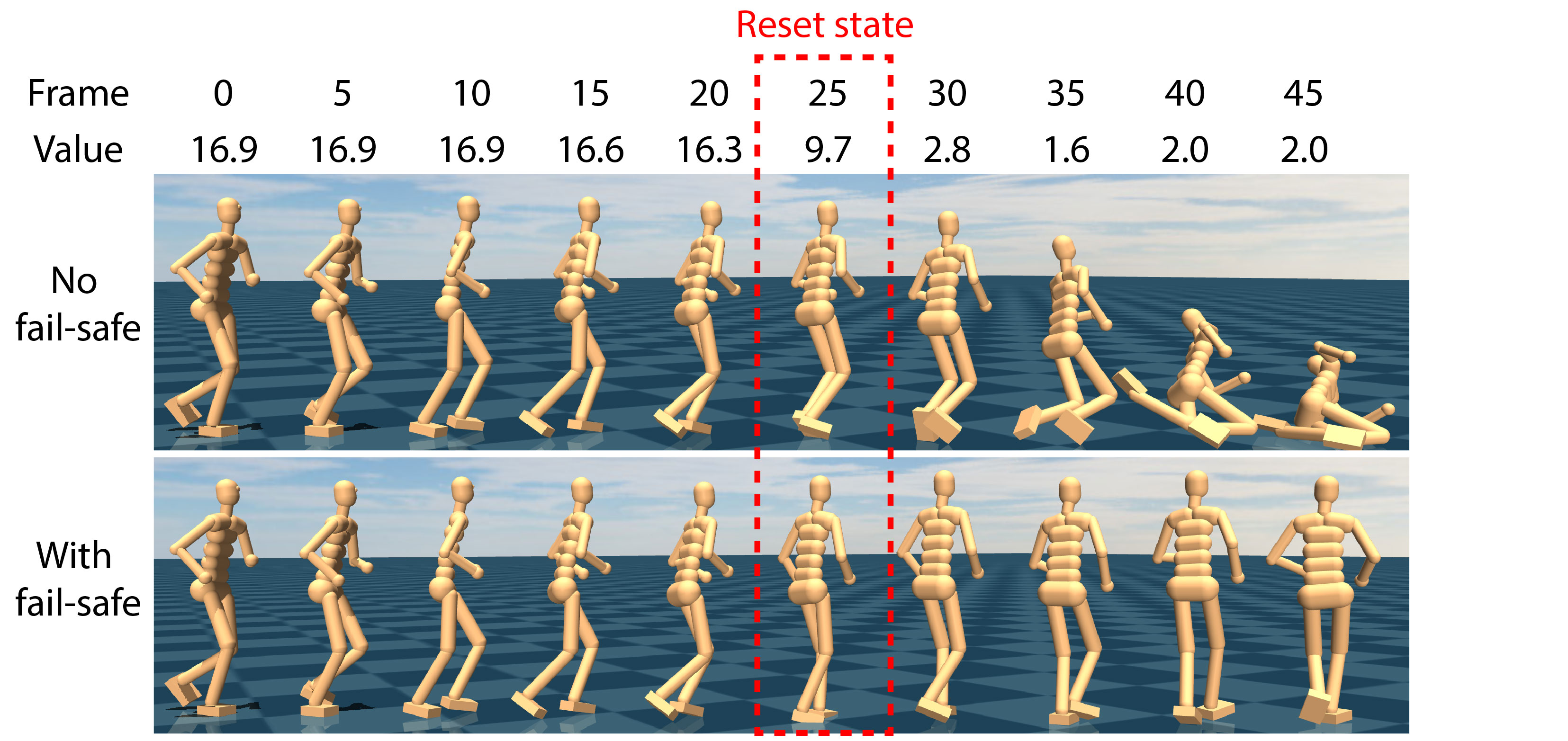}
		\caption{\textbf{Top:} The humanoid at unstable state falls to the ground and the value of the state drops drastically during falling. \textbf{Bottom:} At frame 25, the instability is detected by our fail-safe mechanism, which triggers the state reset and allows our method to keep producing good pose estimates.}
		\label{fig:failsafe}
		\vspace{-5mm}
	\end{figure}
	When running ego-pose estimation at test time, even though the control policy $\pi$ is often robust enough to recover from errors, the humanoid can still fall due to irregular actions caused by extreme domain shifts in the visual input. When the humanoid falls, we need to reset the humanoid state to the output of the state regressor $\mathcal{F}$ to keep producing meaningful pose estimates. However, it is not clear when to do the reset. A naive solution is to reset the state when the humanoid falls to the ground, which will generate a sequence of bad pose estimates during falling (Fig.~\ref{fig:failsafe} (Top)). We propose a fail-safe mechanism that can detect the instability of current state before the humanoid starts to fall, which enables us to reset the state before producing bad estimates (Fig.~\ref{fig:failsafe} (Bottom)). Most policy gradient methods have an actor-critic structure, where they train the policy $\pi$ alongside a value function $\mathcal{V}$ which estimates the expected discounted return of a state $s$:
	\begin{equation}
	\mathcal{V}(s) = \mathbb{E}_{s_1 = s,\,a_t \sim \pi} \left[\sum_{t=1}^T \gamma^{t-1} r_t\right]\,.
	\end{equation}
	Assuming that $1 / (1 - \gamma) \ll T$, and for a well-trained policy, $r_t$ varies little across time steps, the value function can be approximated as
	\begin{equation}
	\mathcal{V}(s)\approx \sum_{t=1}^{\infty} \gamma^{t-1} \bar{r}_s = \frac{1}{1-\gamma}\bar{r}_s\,,
	\end{equation}
	where $\bar{r}_s$ is the average reward received by the policy starting from state $s$. During our experiments, we find that for state $s$ that is stable (not falling), its value $\mathcal{V}(s)$ is always close to $1/(1-\gamma)\bar{r}$ with little variance, where $\bar{r}$ is the average reward inside a training batch. But when the humanoid begins falling, the value starts dropping significantly (Fig.~\ref{fig:failsafe}).
	This discovery leads us to the following fail-safe mechanism: when executing the humanoid policy $\pi$, we keep a running estimate of the average state value $\bar{\mathcal{V}}$ and reset the state when we find the value of current state is below $\kappa \bar{\mathcal{V}}$, where $\kappa$ is a coefficient determining how sensitive this mechanism is to instability. We set $\kappa$ to 0.6 in our experiments.

	\section{Experimental Setup}
	\subsection{Datasets}
	\label{sec:dataset}
	The main dataset we use to test our method is a large MoCap dataset with synchronized egocentric videos. It includes five subjects and is about an hour long. Each subject is asked to wear a head-mounted GoPro camera and perform various complex human motions for multiple takes. The motions consist of walking, jogging, hopping, leaning, turning, bending, rotating, crouching and transitions between these motions. Each take is about one minute long, and we do not segment or label the motions. To further showcase our method's utility, we also collected an in-the-wild dataset where two new subjects are asked to perform similar actions to the MoCap data. It has 24 videos each lasting about 20s. Both indoor and outdoor videos are recorded in different places. Because it is hard to obtain ground-truth 3D poses in real-world environment, we use a third-person camera to capture the side-view of the subject, which is used for evaluation based on 2D keypoints.
	
	\subsection{Baselines}
	For ego-pose estimation, we compare our method against three baselines:
	\begin{itemize}[leftmargin=*]
		\setlength\itemsep{-3pt}
		\item \textbf{VGAIL~\cite{yuan20183d}:} a control-based method that uses joint torques as action space, and learns the control policy with video-conditioned GAIL.
		\item \textbf{PathPose:} an adaptation of a CRF-based method~\cite{jiang2017seeing}. We do not use static scene cues as the training data is from MoCap.
		\item \textbf{PoseReg:} a method that uses our state estimator $\mathcal{F}$ to output the kinematic pose sequence directly. We integrate the linear and angular velocities of the root joint to generate global positions and orientations.
	\end{itemize}
	
	For ego-pose forecasting, no previous work has tried to forecast future human poses from egocentric videos, so we compare our approach to methods that forecast future motions using past poses, which at test time is provided by our ego-pose estimation algorithm:
	\begin{itemize}[leftmargin=*]
		\setlength\itemsep{-3pt}
		\item \textbf{ERD~\cite{fragkiadaki2015recurrent}:} a method that employs an encoder-decoder structure with recurrent layers in the middle, and predicts the next pose using current ground-truth pose as input. It uses noisy input at training to alleviate drift.
		\item \textbf{acLSTM~\cite{li2017auto}:} a method similar to ERD with a different training scheme for more stable long-term prediction: it schedules fixed-length fragments of predicted poses as input to the network.
	\end{itemize}
	
	\subsection{Metrics}
	To evaluate both the accuracy and physical correctness of our approach, we use the following metrics:
	\begin{itemize}[leftmargin=*]
		\setlength\itemsep{-3pt}
		\item \textbf{Pose Error} ($\mathbf{E}_{\textrm{pose}}$): a pose-based metric that measures the Euclidean distance between the generated pose sequence $p_{1:T}$ and the ground-truth pose sequence $\hat{p}_{1:T}$. It is calculated as $\frac{1}{T}\sum_{t=1}^T||p_t - \hat{p}_t||_2$.
		
		\item \textbf{2D Keypoint Error} ($\mathbf{E}_{\textrm{key}}$): a pose-based metric used for our in-the-wild dataset. It can be calculated as $\frac{1}{TJ}\sum_{t=1}^T\sum_{j=1}^J||x_t^j - \hat{x}_t^j||_2$, where $x_t^j$ is the $j$-th 2D keypoint of our generated pose and $\hat{x}_t^j$ is the ground truth extracted with OpenPose~\cite{cao2017realtime}. We obtain 2D keypoints for our generated pose by projecting the 3D joints to an image plane with a side-view camera. For both generated and ground-truth keypoints, we set the hip keypoint as the origin and scale the coordinate to make the height between shoulder and hip equal 0.5.
		
		\item \textbf{Velocity Error} ($\mathbf{E}_{\textrm{vel}}$): a physics-based metric that measures the Euclidean distance between the generated velocity sequence $v_{1:T}$ and the ground-truth $\hat{v}_{1:T}$. It is calculated as $\frac{1}{T}\sum_{t=1}^T||v_t - \hat{v}_t||_2$. $v_t$ and $\hat{v}_t$ can be computed by the finite difference method.
		
		\item \textbf{Average Acceleration} ($\mathbf{A}_{\textrm{accl}}$): a physics-based metric that uses the average magnitude of joint accelerations to measure the smoothness of the generated pose sequence. It is calculated as $\frac{1}{TG}\sum_{t=1}^T||\dot{v}_t||_1$ where $\dot{v}_t$ denotes joint accelerations and $G$ is the number of actuated DoFs. 
		
		\item \textbf{Number of Resets} ($\mathbf{N}_{\textrm{reset}}$): a metric for control-based methods (Ours and VGAIL) to measure how frequently the humanoid becomes unstable.
	\end{itemize}
	
	\subsection{Implementation Details}
	\vspace{1mm}\noindent\textbf{Simulation and Humanoid.}
	We use MuJoCo~\cite{todorov2012mujoco} as the physics simulator.
	The humanoid model is constructed from the BVH file of a single subject and is shared among other subjects. The humanoid consists of 58 DoFs and 21 rigid bodies with proper geometries assigned. Most non-root joints have three DoFs except for knees and ankles with only one DoF. We do not add any stiffness or damping to the joints, but we add 0.01 armature inertia to stabilize the simulation. We use stable PD controllers~\cite{tan2011stable} to compute joint torques. The gains $k_p$ ranges from 50 to 500 where joints such as legs and spine have larger gains while arms and head have smaller gains. Preliminary experiments showed that the method is robust to a wide range of gains values. $k_d$ is set to $0.1k_p$. We set the torque limits based on the gains.
	
	\vspace{1mm}\noindent\textbf{Networks and Training.} For the video context networks, we use PWC-Net~\cite{sun2018pwc} to compute optical flow and ResNet-18~\cite{he2016deep} pretrained on ImageNet to generate the visual features $\psi_t \in \mathbb{R}^{128}$. To accelerate training, we precompute $\psi_t$ for the policy using the ResNet pretrained for initial state estimation. We use a BiLSTM (estimation) or LSTM (forecasting) to produce the visual context $\phi_t \in \mathbb{R}^{128}$. For the policy, we use online z-filtering to normalize humanoid state $z_t$, and the diagonal elements of the covariance matrix $\Sigma$ are set to {$0.1$}. When training for pose estimation, for each episode we randomly sample a data fragment of 200 frames ({6.33}s) and pad 10 frames of visual features $\psi_t$ on both sides to alleviate border effects when computing $\phi_t$. When training for pose forecasting, we sample 120 frames and use the first 30 frames as context to forecast 90 future frames. We terminate the episode if the humanoid falls or the time horizon is reached. For the reward weights $(w_q, w_e, w_p, w_v)$, we set them to (0.5, 0.3, 0.1, 0.1) for estimation and (0.3, 0.5, 0.1, 0.1) for forecasting. We use PPO~\cite{schulman2017proximal} with a clipping epsilon of 0.2 for policy optimization. The discount factor $\gamma$ is 0.95. We collect trajectories of 50k timesteps at each iteration. We use Adam~\cite{kingma2014adam} to optimize the policy and value function with learning rate 5e-5 and 3e-4 respectively. The policy typically converges after 3k iterations, which takes about 2 days on a GTX 1080Ti.

	\begin{figure}
		\centering
		\includegraphics[width=0.95\linewidth]{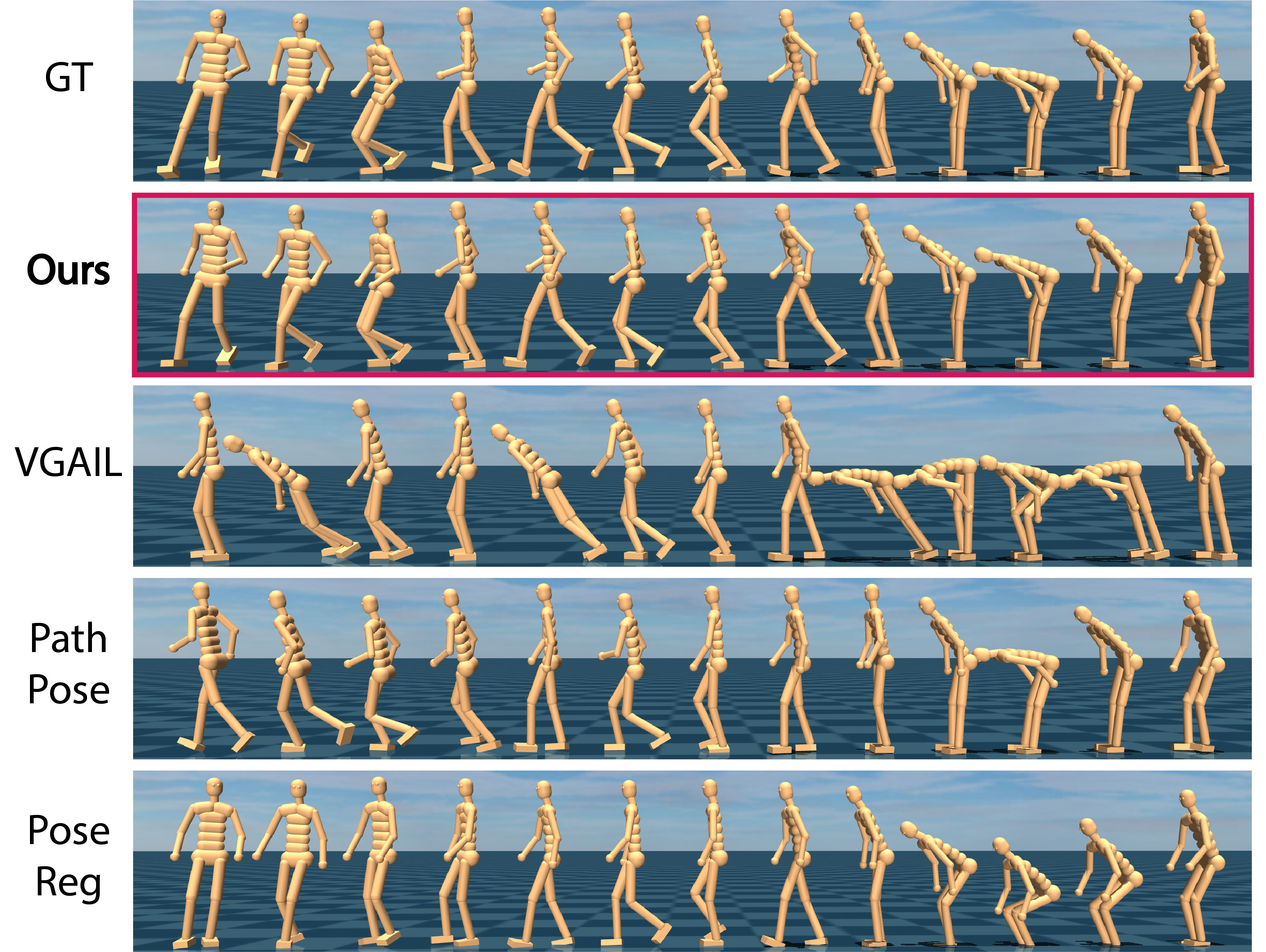}
		\caption{Single-subject ego-pose estimation results.}
		\label{fig:single_estimate}
		\vspace{-3mm}
	\end{figure}
	
	\begin{figure}
		\centering
		\includegraphics[width=0.95\linewidth]{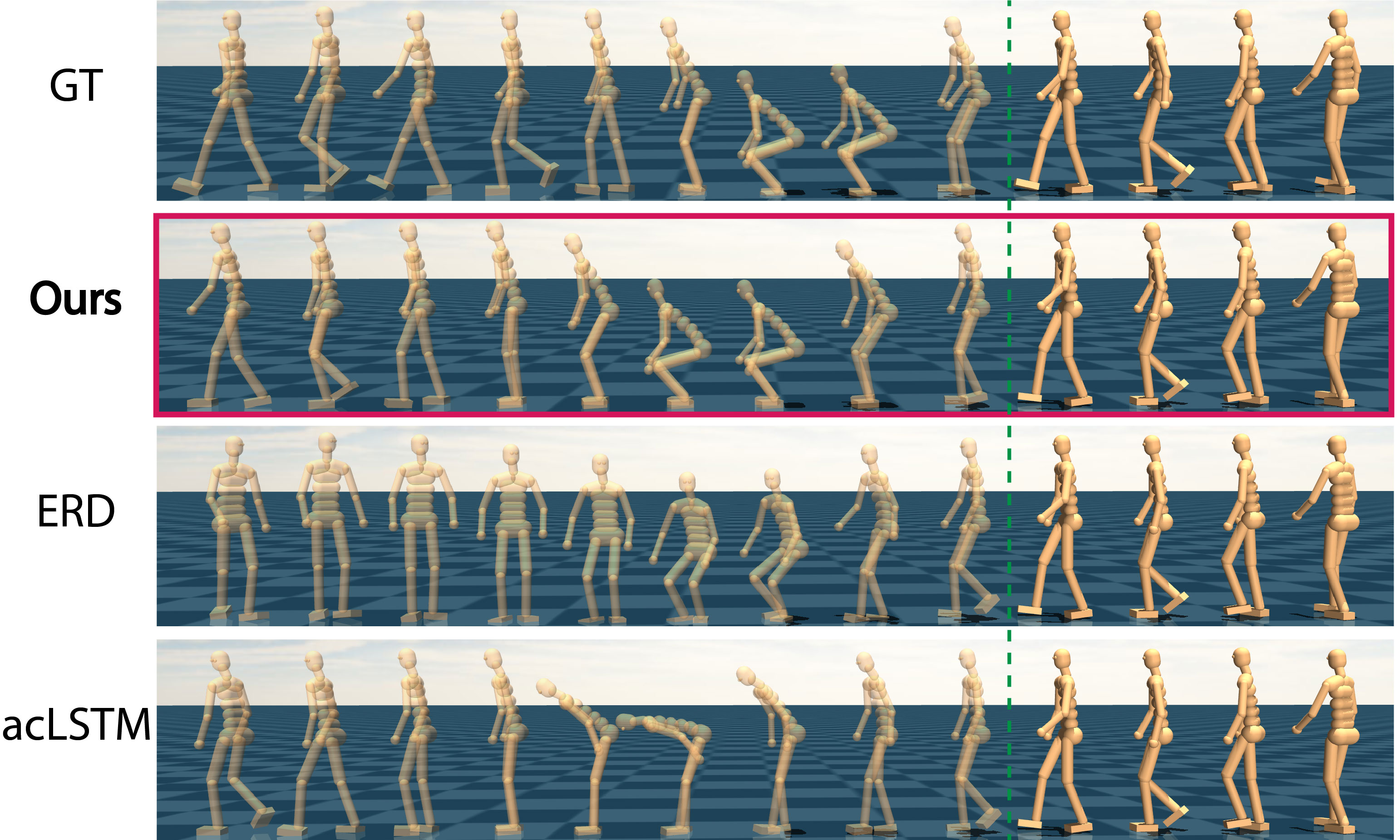}
		\caption{Single-subject ego-pose forecasting results.}
		\label{fig:single_forecast}
		\vspace{-5mm}
	\end{figure}
	
	\begin{figure}
		\centering
		\includegraphics[width=\linewidth]{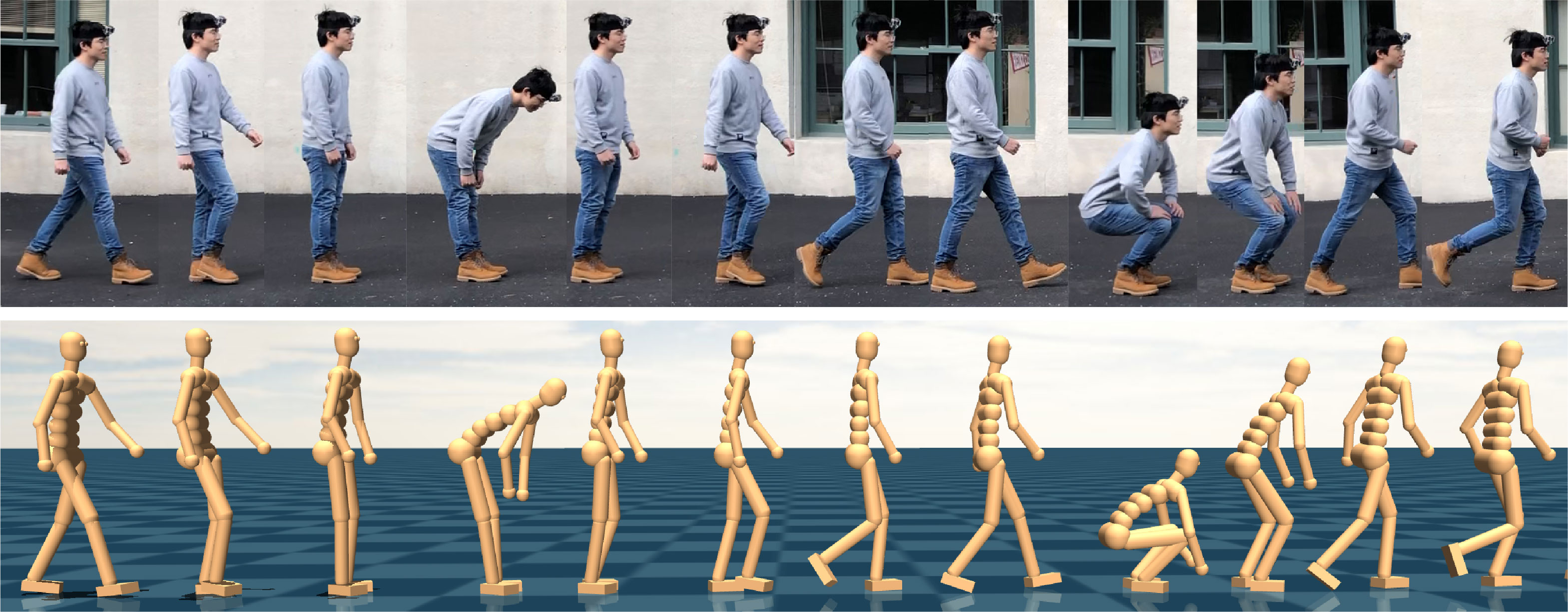}
		\caption{In-the-wild ego-pose estimation results.}
		\label{fig:wild_estimate}
		\vspace{-3mm}
	\end{figure}
	
	\begin{figure}
		\centering
		\includegraphics[width=\linewidth]{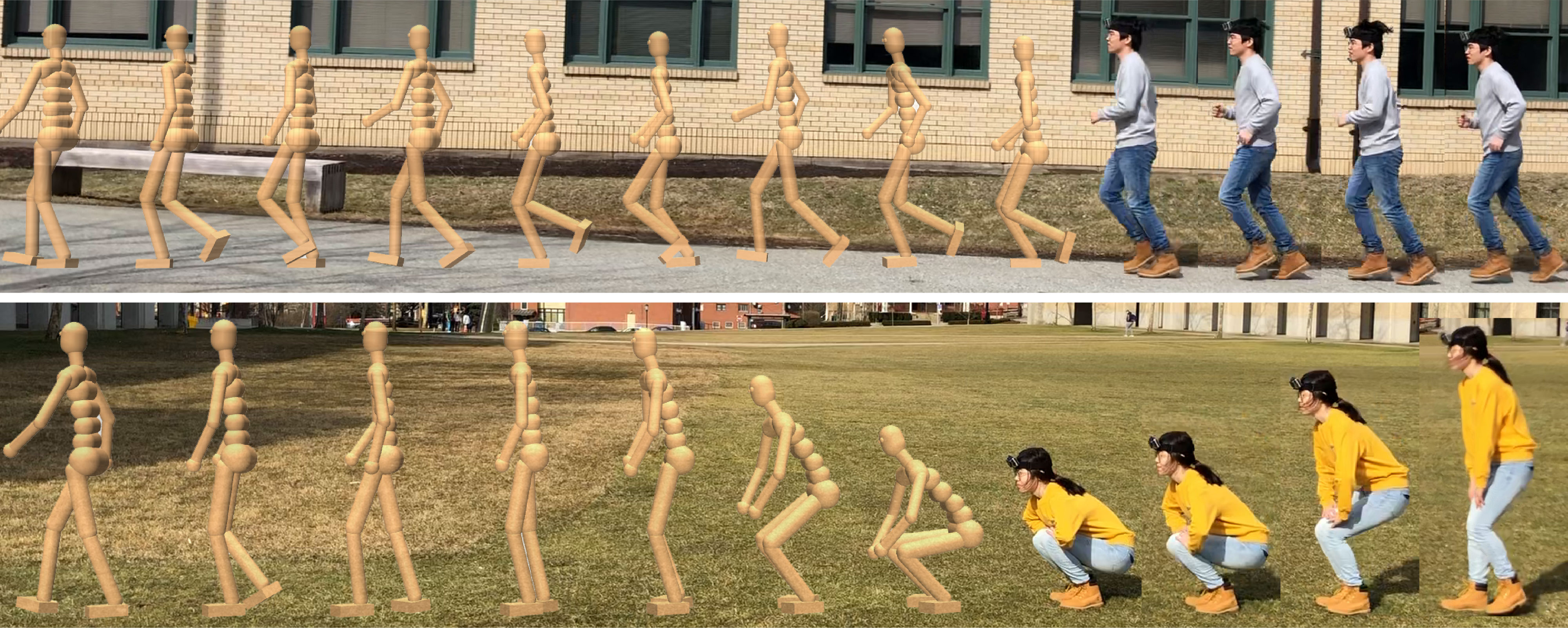}
		\caption{In-the-wild ego-pose forecasting results.}
		\label{fig:wild_forecast}
		\vspace{-5mm}
	\end{figure}

	\section{Results}
	\begin{table*}[h]
		\footnotesize
		\centering
		\begin{tabular}{@{\hskip 2mm}lr@{\hskip -0mm}rrrrrrrrrrrr@{\hskip 2mm}}
			\toprule
			\multicolumn{14}{c}{\textsc{\textbf{Ego-pose Estimation}}}\\ \midrule
			&& \multicolumn{4}{c}{Single Subject} && \multicolumn{4}{c}{Cross Subjects} & & \multicolumn{2}{c}{In the Wild} \\ \cmidrule{3-6} \cmidrule{8-11} \cmidrule{13-14}
			Method && $\mathbf{E}_{\textrm{pose}}$ & $\mathbf{N}_{\textrm{reset}}$ & $\mathbf{E}_{\textrm{vel}}$ & $\mathbf{A}_{\textrm{accl}}$ && $\mathbf{E}_{\textrm{pose}}$ & $\mathbf{N}_{\textrm{reset}}$ & $\mathbf{E}_{\textrm{vel}}$ & $\mathbf{A}_{\textrm{accl}}$ && $\mathbf{E}_{\textrm{key}}$ & $\mathbf{A}_{\textrm{accl}}$ \\ \midrule
			Ours   && \textbf{0.640} & \textbf{1.4} & \textbf{4.469} & \textbf{5.002} && \textbf{1.183} & \textbf{4} & \textbf{5.645} & \textbf{5.260} && \textbf{0.099} & \textbf{5.795}\\  
			VGAIL~\cite{yuan20183d}  && 0.978 & 94 & 6.561 & 9.631 && 1.316 & 418 & 7.198 & 8.837 && 0.175 & 9.278\\
			PathPose~\cite{jiang2017seeing} && 1.035 & -- & 19.135 & 63.526 && 1.637 & -- & 32.454 & 117.499 && 0.147 & 125.406\\
			PoseReg && 0.833 & -- & 5.450 & 7.733 && 1.308 & -- & 6.334 & 8.281 && 0.109 & 7.611 \\
			\midrule
			\midrule
			\multicolumn{14}{c}{\textsc{\textbf{Ego-pose Forecasting}}}\\ \midrule
			&& \multicolumn{4}{c}{Single Subject} && \multicolumn{4}{c}{Cross Subjects} & & \multicolumn{2}{c}{In the Wild} \\ \cmidrule{3-6} \cmidrule{8-11} \cmidrule{13-14}
			Method && $\mathbf{E}_{\textrm{pose}}$ & $\mathbf{E}_{\textrm{pose}}$(3s) & $\mathbf{E}_{\textrm{vel}}$ & $\mathbf{A}_{\textrm{accl}}$ && $\mathbf{E}_{\textrm{pose}}$ & $\mathbf{E}_{\textrm{pose}}$(3s) & $\mathbf{E}_{\textrm{vel}}$ & $\mathbf{A}_{\textrm{accl}}$ & & $\mathbf{E}_{\textrm{key}}$ & $\mathbf{A}_{\textrm{accl}}$ \\ \midrule
			Ours   && \textbf{0.833} & \textbf{1.078} & \textbf{5.456} & \textbf{4.759} && \textbf{1.179} & \textbf{1.339} & \textbf{6.045} & \textbf{4.210} && \textbf{0.114} & \textbf{4.515}\\  
			ERD~\cite{fragkiadaki2015recurrent}  && 0.949 & 1.266 & 6.242 & 5.916 && 1.374 & 1.619 & 7.238 & 6.419 && 0.137 & 7.021\\
			acLSTM~\cite{li2017auto} && 0.861 & 1.232 & 6.010 & 5.855 && 1.314 & 1.511 & 7.454 & 7.123 && 0.134 & 8.177 \\
			\bottomrule
		\end{tabular}
		\vspace{2mm}
		\caption{Quantitative results for egocentric pose estimation and forecasting. For forecasting, by default the metrics are computed inside the first 1s window, except that $\mathbf{E}_{\textrm{pose}}$(3s) are computed in the first 3s window.}
		\label{table:big}
		\vspace{-3mm}
	\end{table*}

	\begin{table}[h]
		\footnotesize
		\centering
		\begin{tabular}{@{\hskip 1mm}lrrrr@{\hskip 1mm}}
			\toprule
			Method & $\mathbf{N}_{\textrm{reset}}$ & $\mathbf{E}_{\textrm{pose}}$ & $\mathbf{E}_{\textrm{vel}}$ & $\mathbf{A}_{\textrm{accl}}$ \\ \midrule
			(a) Ours   & \textbf{4} & \textbf{1.183} & \textbf{5.645} & \textbf{5.260}\\  
			(b) Partial reward $r_q + r_e$ & 55 & 1.211 & 5.730 & 5.515\\
			(c) Partial reward $r_q$  & 14 & 1.236 & 6.468 & 8.167\\
			(d) DeepMimic reward~\cite{peng2018deepmimic} & 52 & 1.515 & 7.413 & 17.504 \\
			(e) No fail-safe & 4 & 1.206 & 5.693 & 5.397 \\
			\bottomrule
		\end{tabular}
		\vspace{1mm}
		\caption{Ablation study for ego-pose estimation.}
		\label{table:ablation_estimate}
		\vspace{-5mm}
	\end{table}
	
	To comprehensively evaluate performance, we test our method against other baselines in three different experiment settings: (1) single subject in MoCap; (2) cross subjects in MoCap; and (3) cross subjects in the wild. We further conduct an extensive ablation study to show the importance of each technical contributon of our approach. Finally, we show time analysis to validate that our approach can run in real-time.
	
	\vspace{1mm}\noindent\textbf{Subject-Specific Evaluation.}
	In this setting, we train an estimation model and a forecasting model for each subject. We use a 80-20 train-test data split. For forecasting, we test every 1s window to forecast poses in the next 3s. The quantitative results are shown in Table~\ref{table:big}. For ego-pose estimation, we can see our approach outperforms other baselines in terms of both pose-based metric (pose error) and physics-based metrics (velocity error, acceleration, number of resets). We find that VGAIL~\cite{yuan20183d} is often unable to learn a stable control policy from the training data due to frequent falling, which results in the high number of resets and large acceleration. For ego-pose forecasting, our method is more accurate than other methods for both short horizons and long horizons. We also present qualitative results in Fig.~\ref{fig:single_estimate} and~\ref{fig:single_forecast}. Our method produces pose estimates and forecasts closer to the ground-truth than any other baseline. 
	
	\vspace{1mm}\noindent\textbf{Cross-Subject Evaluation.} To further test the robustness of our method, we perform cross-subject experiments where we train our models on four subjects and test on the remaining subject. This is a challenging setting since people have very unique style and speed for the same action. As shown in Table~\ref{table:big}, our method again outperforms other baselines in all metrics and is surprisingly stable with only a small number of resets. For forecasting, we also show in Table~\ref{table:forecast_horizon} how pose error changes across different forecasting horizons. We can see our forecasting method is accurate for short horizons ($<1$s) and even achieves comparable results as our pose estimation method (Table~\ref{table:big}).
	
	\vspace{1mm}\noindent\textbf{In-the-Wild Cross-Subject.} To showcase our approach's utility in real-world scenarios, we further test our method on the in-the-wild dataset described in Sec.~\ref{sec:dataset}. Due to the lack of 3D ground truth, we make use of accompanying third-person videos and compute 2D keypoint error as the pose metric. As shown in Table~\ref{table:big}, our approach is more accurate and smooth than other baselines for real-world scenes. We also present qualitative results in Fig.~\ref{fig:wild_estimate} and~\ref{fig:wild_forecast}. For ego-pose estimation (Fig.~\ref{fig:wild_estimate}), our approach produces very accurate poses and the phase of the estimated motion is synchronized with the ground-truth motion. For ego-pose forecasting (Fig.~\ref{fig:wild_forecast}), our method generates very intuitive future motions, as a person jogging will keep jogging forward and a person crouching will stand up and start to walk. 
	
	\vspace{1mm}\noindent\textbf{Ablative Analysis.} The goal of our ablation study is to evaluate the importance of our reward design and fail-safe mechanism. We conduct the study in the cross-subject setting for the task of ego-pose estimation. We can see from Table~\ref{table:ablation_estimate} that using other reward functions will reduce performance in all metrics. We note that the large acceleration in (b) and (c) is due to jittery motions generated from unstable control policies. Furthermore, by comparing (e) to (a) we can see that our fail-safe mechanism can improve performance even though the humanoid seldom becomes unstable (only 4 times).
	
	\vspace{1mm}\noindent\textbf{Time analysis.}
	We perform time analysis on a mainstream CPU with a GTX 1080Ti using PyTorch implementation of ResNet-18 and PWCNet\footnote{https://github.com/NVlabs/PWC-Net}.
	The breakdown of the processing time is: optical flow 5ms, CNN 20ms, LSTM + MLP 0.2ms, simulation 3ms. The total time per step is $\sim30$ms which translates to 30 FPS. To enable real-time pose estimation which uses a bi-directional LSTM, we use a 10-frame look-ahead video buffer and only encode these 10 future frames with our backward LSTM, which corresponds to a fixed latency of 1/3s. For pose forecasting, we use multi-threading and run the simulation on a separate thread. Forecasting is performed every 0.3s to predict motion 3s (90 steps) into the future. To achieve this, we use a batch size of 5 for the optical flow and CNN (cost is 14ms and 70ms with batch size 1).
	
	\begin{table}
		\footnotesize
		\centering
		\begin{tabular}{@{\hskip 1mm}lrrrrrrrrr@{\hskip 1mm}}
			\toprule
			Method & 1/3s & 2/3s & 1s & 2s & 3s \\ \midrule
			Ours   & \textbf{1.140} & \textbf{1.154} & \textbf{1.179} & \textbf{1.268} & \textbf{1.339}\\  
			ERD~\cite{fragkiadaki2015recurrent} & 1.239 & 1.309 & 1.374 & 1.521 & 1.619 \\
			acLSTM~\cite{li2017auto}  & 1.299 & 1.297 & 1.314 & 1.425 & 1.511 \\
			\bottomrule
		\end{tabular}
		\vspace{1mm}
		\caption{Cross-subject $\mathbf{E}_{\textrm{pose}}$ for different forecasting horizons.}
		\label{table:forecast_horizon}
		\vspace{-5mm}
	\end{table}
	
	\section{Conclusion}
	We have proposed the first approach to use egocentric videos to both estimate and forecast 3D human poses. Through the use of a PD control based policy and a reward function tailored to unsegmented human motion data, we showed that our method can estimate and forecast accurate poses for various complex human motions. Experiments and time analysis showed that our approach is robust enough to transfer directly to real-world scenarios and can run in real-time.
	
	\noindent\textbf{Acknowledgment.} This work was sponsored in part by JST CREST (JPMJCR14E1) and IARPA (D17PC00340).
	
	{\small
		\bibliographystyle{ieee}
		\bibliography{reference}
	}
	
\end{document}